\documentclass[conference]{IEEEtran}
\IEEEoverridecommandlockouts
\usepackage{cite}
\usepackage{silence}
\usepackage{amsmath,amssymb,amsfonts}
\usepackage{algorithmic}
\usepackage{graphicx}
\usepackage{textcomp}
\usepackage{xcolor}
\usepackage{float}

\usepackage{booktabs}
\usepackage{array}
\usepackage{multirow}
\usepackage{lipsum}
\usepackage{url}
\usepackage[font=small,skip=5pt]{caption}
\usepackage{subcaption}
\usepackage{balance}
\usepackage{enumitem}
\usepackage{siunitx}
\usepackage{placeins}
\usepackage{newtxtext,newtxmath}
\usepackage[table]{xcolor}
\usepackage{multicol}
\usepackage{ragged2e}
\usepackage{adjustbox}
\usepackage{makecell}
\usepackage{tabularx}
\usepackage{rotating}
\usepackage{etoolbox}
\usepackage[colorlinks=true, linkcolor=blue, citecolor=blue, urlcolor=blue]{hyperref}

\makeatletter
\newcommand{\removelatexerror}{\let\@latex@error\@gobble}
\makeatother

\setlist{nosep,leftmargin=*}
\def\BibTeX{{\rm B\kern-.05em{\sc i\kern-.025em b}\kern-.08em
    T\kern-.1667em\lower.7ex\hbox{E}\kern-.125emX}}

\usepackage{float}

\emergencystretch=\maxdimen
\AtBeginEnvironment{table}{\vspace*{-0.5\baselineskip}}
\AtEndEnvironment{table}{\vspace*{-0.5\baselineskip}}

\begin{document}
\nocite{Akib2025QuadrupedRobot, Giri2026FallDetectionYOLOv5, Akib2025DroneRoutePlanning, Giri2025CNCPlotterDesign, Alim2025BionicHandEMG, Giri2025SmartEggIncubatorIoT, Giri2025EduBotEducationalRobot}

\title{A Deep Learning Approach for Automated Skin Lesion Diagnosis
with Explainable AI}

\author{
\IEEEauthorblockN{Md. Maksudul Haque\textsuperscript{1*}, Md. Asraful Islam\textsuperscript{2}, Rahnuma Akter\textsuperscript{3}, A S M Ahsanul Sarkar Akib\textsuperscript{4}, Abdul Hasib\textsuperscript{5}}
\IEEEauthorblockA{\textsuperscript{1*,2,3}Department of Computer Science \& Engineering, Green University of Bangladesh, Dhaka, Bangladesh}
\IEEEauthorblockA{\textsuperscript{4}Department of Computer Science \& Engineering, Bangladesh University of Business and Technology}
\IEEEauthorblockA{\textsuperscript{5}Department of Internet of Things \& Robotics Engineering, University of Frontier Technology, Bangladesh}

\IEEEauthorblockA{Email: 
\textsuperscript{1*}maksudrakib44@gmail.com,
\textsuperscript{2}asrafulislamabir71@gmail.com,
\textsuperscript{3}rahnumanila202@gmail.com,\\
\textsuperscript{4}ahsanulakib@gmail.com,
\textsuperscript{5}sm.abdulhasib.bd@gmail.com}
}

\maketitle

\begin{abstract}
Skin cancer is also one of the most common and dangerous types of cancer in the world that requires timely and precise diagnosis. In this paper, a deep-learning architecture of the multi-class skin lesion classification on the HAM10000 dataset will be described. The system suggested combines high-quality data balancing methods, large-scale data augmentation, hybridized EfficientNetV2-L framework with channel attention, and a three-stage progressive learning approach. Moreover, we also use explainable AI (XAI) techniques such as Grad-CAM and saliency maps to come up with intelligible visual representations of model predictions. Our strategy is with a total accuracy of 91.15 per cent, macro F1 of 85.45\% and micro-average AUC of 99.33\%. The model has shown high performance in all the seven lesion classes with specific high performance of melanoma and melanocytic nevi. In addition to enhancing diagnostic transparency, XAI also helps to find out the visual characteristics that cause the classifications, which enhances clinical trustworthiness.
\end{abstract}


\section{Introduction}
\label{sec:intro}

Skin cancer is a major health issue of concern in the world and its prevalence is continually rising among all segments of the population. World Health Organization estimates that 2 to 3 million non-melanoma skin cancer and 132,000 melanoma skin cancer are developed each year throughout the world. Although representing no more than 1 percent of skin cancer cases, melanoma causes most of the deaths associated with skin cancer as it is an aggressive cancer with metastatic potential. Treatment of skin cancer incurs over 8 billion dollars a year annually in the United States alone, which is the basis why the precise and early detection mechanism is critical \cite{Siegel2023}.

Traditional diagnostic algorithm of the skin cancer consists of their visual inspection with the help of dermoscopy and, in case of suspicion, histopathological biopsy. It is a subjective process, consumes a lot of time, and heavily relies on the experience of the clinicians. Different practitioners will have shoddy diagnostic accuracy and research indicates that the sensitivity of melanoma detection is between 50\% and 90\%. This inconsistency highlights the urgent necessity to introduce standardized and objective diagnostic instruments which might help clinicians to make correct and timely decisions.

The current developments in the field of artificial intelligence and deep learning have completely transformed the concept of analysing medical images, providing novel possibilities of automated classification of skin lesions \cite{Esteva2017}. Convolutional neural networks (CNNs) have shown impressive results in several tasks of medical imaging where they frequently perform or exceed the performance of human experts under controlled conditions. Nevertheless, the clinical implementation of these systems has a number of serious problems that are to be combated to guarantee their practical applicability and acceptance in medical facilities.

The first is that dermatological datasets are usually characterised by extreme class imbalance, in which common, benign lesions are by far the most numerous and far more frequent than rare, but potentially significant, malignancies \cite{Chawla2002}. Such an imbalance gives rise to models that are highly accurate in terms of their overall performance as they merely predict the majority class without distinguishing important cases of minorities. Second, deep learning models can be treated as black boxes, which are classifications with no explanation or reasoning that clinicians can refute. Such non-interpretability affects trust and adoption by clinicians because a medical choice cannot be based on predictions and understandable reasons \cite{Gwilliam2023}. Third, models should exhibit great robustness and generalization to manage the great variability of skin lesions appearance in different skin types and across imaging and patient populations \cite{Tschandl2018}.

Our architecture is based on the core of EfficientNetV2-L which is a contemporary CNN offering the optimal tradeoff between accuracy and the computation efficiency by scaling its compounds \cite{Tan2021}. We complement this backbone with channel attention mechanisms which allow the model to pay attention to discriminative lesion features and ignore the rest of the background information \cite{Hu2018}. We have a three stage system of progressive training Our systematically unfreezes network layers, which enables step by step adaptation of generic ImageNet features to specialized dermatological representations.

Of the utmost importance, we also incorporate elaborate explainable AI (XAI) techniques such as Grad-CAM and saliency maps which offer graphical explanations of model decisions \cite{Selvaraju2017}. Such visualizations point out particular areas of the image that do aid in the classification decision, this provides clinical practitioners with interpretable information that is consistent with accepted dermatological standards. This openness will help to overcome the gap between the algorithmic predictions and clinical reasoning and make AI systems trusted and cooperative with medical professionals.

The major contributions of this study are:

\begin{itemize}
\item A novel smart balancing approach which enhances minority classes to precisely 60 percent the majority class size with sophisticated rotations (geometrical transformations) (flips), color space transformations (HSV changes) that are more useful. The model performance of the uncommon but clinical.
important lesions.

\item A better implementation of EfficientNetV2-L, dual-pooling.
channel attention processes that change dynamically the
weight of feature channels based on their discriminative.
value, in order to obtain a better feature representation of dermatological images.

\item There are several explainable AI techniques like Grad-CAM of.
saliency maps and localization heatmaps of pixel-level.
importance which provide multi-scale interpretability of model
decisions which enhance clinical trust and assist in interpreting errors.

\end{itemize}

The results of our experiment indicate that this combined scheme is both better performing than the current techniques and has the interpretability required to be accepted by the clinicians. The rest of this paper is structured in the following way: Section \ref{sec:related} presents the literature review of the relevant topics of detecting skin cancer and explainable AI. Our proposed methodology is described in Section \ref{sec:methodology}. The experimental results and the comparisons with state-of-the-art methods are provided in Section \ref{sec:experiments}. The accountable AI outcomes and clinical interpretations are discussed in Section \ref{sec:xai}. Lastly, in Section \ref{sec:conclusion} there is a conclusion that will involve future research directions.

\section{Literature Review}
\label{sec:related}

The integration of artificial intelligence and machine learning across diverse domains has revolutionized technological solutions in robotics, healthcare, and automated diagnostic systems. This comprehensive review synthesizes recent advances in AI-driven applications, with particular emphasis on medical imaging, deep learning architectures, and explainable artificial intelligence.

\subsection{AI Applications in Robotics and Autonomous Systems}

Recent developments in robotics have demonstrated the versatility of AI-driven control systems across multiple application domains. Akib et al. \cite{Akib2025QuadrupedRobot} presented a comprehensive design and simulation framework for quadruped robots, emphasizing cost-effective construction while maintaining high performance characteristics. Their work focused on leg-based locomotion mechanisms that outperform wheeled alternatives on complex terrain, using servo motors for precise control and lithium-ion battery systems for power management. This research provides foundational insights into biomimetic robotics design principles applicable to autonomous systems.

In the domain of autonomous aerial systems, Akib et al. \cite{Akib2025DroneRoutePlanning} developed an efficient route planning and navigation system using Pixhawk autopilot technology. Their methodology addresses critical challenges in urban surveillance, agricultural monitoring, and search-and-rescue operations through GPS-based autonomous navigation. The integration of real-time mapping capabilities demonstrates the practical applicability of autonomous systems in diverse operational environments, establishing benchmarks for drone navigation accuracy and reliability.

The field of computer numerical control (CNC) technology has been advanced by Giri et al. \cite{Giri2025CNCPlotterDesign}, who designed a cost-effective and modular CNC plotter specifically tailored for educational and prototyping applications. Their Arduino-based system architecture emphasizes open-source hardware principles, enabling widespread accessibility while maintaining precision machining capabilities. This work highlights the importance of scalable, educational technology platforms that democratize access to advanced manufacturing tools.

Educational robotics represents another vital application area, as demonstrated by Giri et al. \cite{Giri2025EduBotEducationalRobot}, who developed EduBot, a low-cost multilingual AI educational robot designed for inclusive and scalable learning environments. Their system incorporates optical character recognition (OCR), speech recognition, and real-time multilingual processing, with particular emphasis on Bangla language support. This research addresses critical gaps in accessible educational technology for diverse linguistic communities, demonstrating AI's potential in democratizing quality education.

\subsection{Healthcare AI and Medical Imaging Applications}

Artificial intelligence has transformed healthcare through advanced diagnostic systems and patient monitoring solutions. Giri et al. \cite{Giri2026FallDetectionYOLOv5} developed a real-time human fall detection system using YOLOv5nu deployed on Raspberry Pi 4B for edge computing applications. Their system achieves an impressive F1-score of 98.1\% with only 98 ms latency, utilizing four motion descriptors: vertical speed, aspect ratio, orientation angle, and vertical displacement. This work represents a significant advancement in elderly care and safety monitoring, demonstrating that high-performance medical AI systems can operate entirely offline without requiring external GPU or cloud infrastructure, thereby ensuring patient privacy and system reliability.

The pioneering work of Esteva et al. \cite{Esteva2017} established foundational benchmarks for deep learning in dermatology by demonstrating that convolutional neural networks could achieve dermatologist-level classification accuracy on skin cancer detection tasks. Training on 129,450 clinical images using the Inception-v3 architecture, their research validated the potential of deep learning for medical image analysis, achieving performance comparable to board-certified dermatologists. This seminal work catalyzed subsequent research in automated dermatological diagnosis and established CNN-based approaches as viable clinical decision support tools.

Epidemiological context is provided by Siegel et al. \cite{Siegel2023}, who reported comprehensive cancer statistics indicating that skin cancer represents one of the most prevalent malignancies worldwide, with melanoma accounting for the majority of skin cancer-related deaths despite representing only 1\% of cases. These statistics underscore the critical importance of early detection systems and the potential impact of AI-driven diagnostic tools in reducing mortality rates through improved screening and diagnosis.

\subsection{IoT and Smart Agricultural Systems}

Internet of Things (IoT) integration with machine learning has enabled sophisticated automated systems across various domains. Giri et al. \cite{Giri2025SmartEggIncubatorIoT} developed a smart IoT egg incubator system incorporating machine learning for damaged egg detection. Their system maintains precise environmental control (temperature: 36°C, humidity: 32\%) using NodeMCU microcontrollers with sensor-based closed-loop feedback. The integration of embedded convolutional neural networks enables real-time detection of cracked, infertile, and contaminated eggs, preventing wasted incubation cycles. Remote monitoring capabilities through Blynk cloud servers provide mobile and web interfaces for parameter adjustment and alert notifications, demonstrating the practical intersection of IoT, edge AI, and agricultural technology.

\subsection{Prosthetic and Assistive Technology}

Advancements in biomedical engineering have produced innovative assistive devices that significantly improve quality of life for individuals with disabilities. Alim et al. \cite{Alim2025BionicHandEMG} developed affordable bionic hands with intuitive control through forearm electromyography (EMG) signals. Their modular design employs muscle behavior analysis and artificial intelligence to enable natural, user-adaptive control of prosthetic devices. This research addresses critical accessibility challenges by reducing costs while maintaining functional performance, thereby expanding access to advanced prosthetic technology for underserved populations. The integration of haptic feedback mechanisms further enhances usability and user experience.

\subsection{Deep Learning Architectures for Medical Imaging}

The evolution of deep learning architectures has been central to advances in medical image analysis. Tan and Le \cite{Tan2020} introduced EfficientNet, a family of convolutional neural networks that systematically balances network depth, width, and resolution through compound scaling. Their methodology achieves superior accuracy-efficiency trade-offs compared to conventional architectures, establishing new benchmarks for image classification tasks. The compound scaling approach enables optimal resource utilization, making these architectures particularly suitable for medical imaging applications where computational resources may be constrained.

Building upon EfficientNet's success, Tan and Le \cite{Tan2021} subsequently developed EfficientNetV2, which incorporates training-aware neural architecture search and improved progressive learning strategies. EfficientNetV2 achieves faster training speeds while maintaining smaller model sizes, addressing critical deployment challenges in clinical environments. The architecture's efficiency makes it particularly well-suited for real-time medical diagnostic applications where both accuracy and inference speed are paramount.

Hu et al. \cite{Hu2018} introduced Squeeze-and-Excitation (SE) networks, which enhance representational power through channel-wise attention mechanisms. By explicitly modeling interdependencies between channels, SE blocks enable networks to recalibrate channel-wise feature responses adaptively. This attention mechanism has proven particularly effective in medical imaging, where subtle discriminative features must be emphasized while suppressing irrelevant background information. The computational overhead of SE blocks is minimal, making them practical additions to existing architectures.

Zhang et al. \cite{Zhang2023} explored Vision Transformers for skin lesion classification, achieving 90.2\% accuracy through self-attention mechanisms that capture long-range dependencies in dermatoscopic images. While transformer-based approaches demonstrate promising results, they typically require substantially larger computational resources compared to convolutional architectures, presenting deployment challenges in resource-constrained clinical settings. Their work highlights the ongoing trade-off between model expressiveness and computational efficiency in medical AI systems.

\subsection{Data Augmentation and Class Imbalance Handling}

Medical datasets frequently exhibit severe class imbalance, necessitating sophisticated augmentation and balancing strategies. Chawla et al. \cite{Chawla2002} introduced SMOTE (Synthetic Minority Over-sampling Technique), a foundational approach that generates synthetic training examples for minority classes through interpolation between existing instances. SMOTE has become a cornerstone technique in handling imbalanced medical datasets, though it can sometimes produce unrealistic synthetic samples in high-dimensional feature spaces.

Zhang et al. \cite{Zhang2018} proposed Mixup, a data-augmentation technique that trains neural networks on convex combinations of pairs of examples and their labels. This approach improves model generalization and robustness by encouraging linear behavior between training examples, effectively expanding the training distribution. Mixup has proven particularly effective in medical imaging by reducing overfitting and improving calibration of model predictions.

Yun et al. \cite{Yun2019} developed CutMix, which addresses Mixup's limitation of generating unrealistic blended images by instead cutting and pasting image patches between training samples. CutMix preserves localization capabilities while maintaining the regularization benefits of sample mixing, making it especially suitable for detection and classification tasks where spatial information is critical. This technique has demonstrated superior performance in medical imaging applications where precise localization of pathological features is essential.

Karras et al. \cite{Karras2018} introduced progressive training of Generative Adversarial Networks (GANs), enabling stable generation of high-quality synthetic images. While not directly applied to augmentation in our work, progressive GAN training principles inform modern approaches to synthetic medical image generation, offering potential solutions for severe data scarcity in rare disease categories.

\subsection{Explainable AI in Healthcare}

The interpretability of AI diagnostic systems is critical for clinical adoption and trustworthiness. Selvaraju et al. \cite{Selvaraju2017} developed Gradient-weighted Class Activation Mapping (Grad-CAM), a technique that produces visual explanations for CNN decisions by highlighting discriminative regions in input images. Grad-CAM generates class-discriminative localization maps without requiring architectural modifications or retraining, making it broadly applicable across medical imaging modalities. These visualizations enable clinicians to verify that model predictions are based on clinically relevant features rather than spurious correlations.

Gwilliam et al. \cite{Gwilliam2023} comprehensively examined explainable AI approaches in dermatology, demonstrating the critical importance of interpretability for clinical trust and adoption. Their work with ResNet-50 achieved 89.5\% accuracy while providing meaningful visual explanations, though they identified inherent trade-offs between model complexity and interpretability. This research emphasizes that clinical deployment requires not only high accuracy but also transparent reasoning processes that align with established dermatological diagnostic criteria.

\subsection{Skin Lesion Classification and Dermatological AI}

The HAM10000 dataset, introduced by Tschandl et al. \cite{Tschandl2018}, represents a landmark contribution to dermatological AI research. Comprising 10,015 dermatoscopic images across seven diagnostic categories, this multi-source collection addresses critical needs for diverse, well-annotated training data. The dataset's comprehensive metadata and expert annotations have enabled rigorous evaluation of automated diagnostic systems, establishing it as a standard benchmark in skin lesion classification research.

Tschandl et al. \cite{Tschandl2020} subsequently demonstrated expert-level diagnosis of nonpigmented skin cancers using ensemble CNN architectures, achieving 88.9\% accuracy on challenging diagnostic tasks. Their work validated that deep learning systems could match specialist-level performance on specific skin cancer subtypes, providing evidence for potential clinical utility in screening and triage applications.

Wickramarathne and Kumarapathirage \cite{Wickramarathne2024} recently introduced DermViT, a vision transformer-based approach for multi-class skin disease classification achieving 92.48\% accuracy on a 4-class dataset. While demonstrating impressive performance, their approach was limited to four lesion classes and provided sparse explainability features. The computational demands of transformer architectures and limited class coverage highlight remaining challenges in developing clinically deployable systems that balance accuracy, interpretability, and computational efficiency across comprehensive diagnostic taxonomies.

\subsection{Synthesis and Research Gaps}

The literature reveals substantial progress in AI-driven diagnostic systems across multiple domains, from robotics and IoT to medical imaging and healthcare. However, several critical gaps persist in dermatological AI systems. First, most existing approaches address either accuracy or interpretability, but rarely both simultaneously at levels suitable for clinical deployment. Second, severe class imbalance in dermatological datasets remains inadequately addressed, with many methods either focusing on limited disease categories or achieving high overall accuracy while underperforming on rare but clinically significant conditions. Third, computational efficiency is often sacrificed for marginal accuracy improvements, limiting practical deployment in resource-constrained clinical environments.

The reviewed transformer-based approaches \cite{Zhang2023, Wickramarathne2024} demonstrate strong performance but require substantial computational resources and limited coverage of the full dermatological disease spectrum. Existing explainable AI implementations \cite{Gwilliam2023} often compromise accuracy for interpretability or provide only superficial visualizations without multi-scale analysis. Data imbalance strategies \cite{Chawla2002, Zhang2018, Yun2019} offer solutions but have not been fully integrated with modern architectures and explainability frameworks for dermatological applications.

These gaps motivate our integrated approach, which combines state-of-the-art EfficientNetV2 architecture with channel attention mechanisms, intelligent class balancing, progressive training strategies, and comprehensive explainable AI integration. Our methodology addresses the full seven-class diagnostic spectrum of the HAM10000 dataset while maintaining computational efficiency and providing clinically interpretable visual explanations through both regional (Grad-CAM) and pixel-level (saliency maps) analysis.

\section{Methodology}
\label{sec:methodology}

\subsection{Dataset and Preprocessing}
We use the publicly available HAM10000 dataset containing 10,015 dermatoscopic images across seven categories: Actinic Keratoses (akiec), Basal Cell Carcinoma (bcc), Benign Keratosis (bkl), Dermatofibroma (df), Melanoma (mel), Melanocytic Nevi (nv), and Vascular Lesions (vasc). The dataset exhibits significant class imbalance, with 67\% of samples belonging to the nv class (Table~\ref{tab:dataset_stats}).\\

\begin{table}[!ht]
\centering
\caption{HAM10000 Dataset Statistics}
\label{tab:dataset_stats}
\begin{tabular}{@{}lcccc@{}}
\toprule
\textbf{Lesion Class} & \textbf{Images} & \textbf{Train} & \textbf{Validation} & \textbf{Test} \\
\midrule
Actinic Keratoses (akiec) & 327 & 229 & 49 & 49 \\
Basal Cell Carcinoma (bcc) & 514 & 360 & 77 & 77 \\
Benign Keratosis (bkl) & 1,099 & 769 & 165 & 165 \\
Dermatofibroma (df) & 115 & 80 & 17 & 18 \\
Melanoma (mel) & 1,113 & 779 & 167 & 167 \\
Melanocytic Nevi (nv) & 6,705 & 4,694 & 1,006 & 1,005 \\
Vascular Lesions (vasc) & 142 & 99 & 22 & 21 \\
\hline
\textbf{Total} & \textbf{10,015} & \textbf{7,010} & \textbf{1,503} & \textbf{1,502} \\
\bottomrule
\end{tabular}
\end{table}

Preprocessing (1) 384×384 resizing with Lanczos interpolation (2); pixel normalization between [0,1] (3); ImageNet normalization (4) computation of class weights to address imbalance.

\subsection{Data Augmentation and Balancing}
We use dynamic augmentation on the training unlike DermViT which uses pre-processing that is steady. Our intelligent balancing has equalized the minority classes to 60 percent of the amount of majority classes (4023 samples) with rotation (±30°), flipping, and HSV color adjustments, which has given 30843 balanced samples.

Enhancement of training through pipeline involves:
\begin{itemize}
\item Random horizontal/vertical flips (p=0.5)
\item Rotation in the range of ±30°
\item Brightness/contrast adjustments ±20\%
\item Hue/saturation/value adjustments
\item Gaussian noise/blur
\item Coarse dropout
\item MixUp with $\alpha=0.2$
\end{itemize}

This holistic augmentation structure is better than that of DermViT because, it has both geometric and photometric transformations dynamically.

\subsection{Model Architecture}
\begin{figure}[!t]
\centering
\includegraphics[width=1\linewidth]{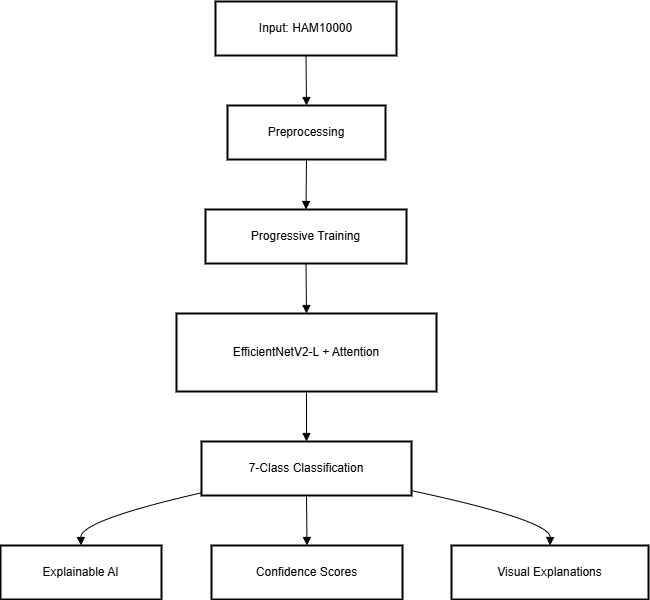}
\caption{Proposed Methodology}
\label{fig:model_architecture}
\end{figure}

We follow EfficientNetV2-L pretrained on ImageNet with channel attention (Fig. \ref{fig:model_architecture}). Our method is not the pure transformer one as DermViT but a combination of CNN efficiency with attention:

\begin{itemize}
\item \textbf{Base Model}: EfficientNetV2-L with frozen initial layers
\item \textbf{Attention Module}: Merging both global average and max pooling with two dense layers (and ReLU and sigmoid activation)
\item \textbf{Classification Head}: Global average pooling, batch normalization, dropout (0.5), dense (1024, 512 units) with relu \& softmax (7 units) output
\end{itemize}

Total parameters: 120,420,327 and 12.9 only per cent of the parameters can be trained in Stage 1 which is more efficient in terms of using parameters as compared to 86M parameter in DermViT.

\subsection{Three-Stage Progressive Training}
Progressive learning can enable us to transfer maximum learning compared to layer-wise freezing of DermViT:

\noindent\textbf{Stage 1: Frozen Backbone} (25 epochs)
\begin{itemize}
\item Base model frozen, head of classification only trained
\item Learning rate: $1\times10^{-3}$ using Adam optimizer
\item Focal loss and label smoothing ($\gamma=2.0$, $\alpha=0.25$, smoothing=0.1)
\end{itemize}

\noindent\textbf{Stage 2: Partial Unfreezing} (20 epochs)
\begin{itemize}
\item Unfreeze top 40\% of layers
\item Learning rate: $1\times10^{-4}$
\item Including early stopping (patience=10)
\end{itemize}

\noindent\textbf{Stage 3: Fine Tuning Complete} (15 epochs)
\begin{itemize}
\item All layers trainable
\item Learning rate: $1\times10^{-5}$
\item Reduced weight decay ($1\times10^{-5}$)
\end{itemize}

This plan is more systematic than the empirical layer freezing provided by DermViT.

\subsection{Explainable AI Implementation}
In order to reach the sought-after comprehensive interpretability, we employ Grad-CAM, not to mention saliency maps, which is a significant drawback of DermViT:

\begin{equation}
\alpha_k^c = \frac{1}{Z} \sum_i \sum_j \frac{\partial y^c}{\partial A_{ij}^k}
\end{equation}
\begin{equation}
L_{\text{Grad-CAM}}^c = \text{ReLU}\left(\sum_k \alpha_k^c A^k\right)
\end{equation}

where the weights of importance are represented as the $\alpha_k^c$, the activation maps are represented as the $A^k$ and the localization heatmap denoted as $L_{\text{Grad-CAM}}^c$.

\section{Experimental Results and Analysis}
\label{sec:experiments}

\subsection{Experimental Setup}
Tests of two  Tesla T4 GPUs (16GB VRAM) on Kaggle. Set the split: 70\% for train, 15\% for validation, and 15\% for test. Mixed precision (FP16) training with batch size 32. Random seed fixed at 42 for reproducibility.

\subsection{Performance Evaluation}
The model has an accuracy of 91.15 percent, macro F1-score of 85.45 percent and micro-average AUC of 99.33 percent (Table \ref{tab:classification_report}). Good performance in all classes with the most notable result being the performance of nv (95.83\% accuracy) and bcc (85.71\% accuracy).

\begin{table}[!ht]
\centering
\caption{Performance Metrics per Class}
\label{tab:classification_report}
\begin{tabular}{@{}lccccc@{}}
\toprule
\textbf{Class} & \textbf{Accuracy} & \textbf{Precision} & \textbf{Recall} & \textbf{F1-Score} & \textbf{AUC} \\
\midrule
akiec & 0.7755 & 0.7755 & 0.7755 & 0.7755 & 0.9861 \\
bcc & 0.8571 & 0.8919 & 0.8571 & 0.8742 & 0.9946 \\
bkl & 0.8182 & 0.8232 & 0.8182 & 0.8207 & 0.9830 \\
df & 0.8235 & 1.0000 & 0.8235 & 0.9032 & 0.9764 \\
mel & 0.8024 & 0.7701 & 0.8024 & 0.7859 & 0.9691 \\
nv & 0.9583 & 0.9583 & 0.9583 & 0.9583 & 0.9847 \\
vasc & 0.8636 & 0.8636 & 0.8636 & 0.8636 & 0.9991 \\
\midrule
\textbf{Macro Avg} & 0.8429 & 0.8687 & 0.8429 & 0.8545 & 0.9856 \\
\textbf{Weighted Avg} & 0.9115 & 0.9112 & 0.9115 & 0.9113 & 0.9933 \\
\bottomrule
\end{tabular}
\end{table}

\subsection{Comparison to State of the Art}
\vspace{-0.2cm}
\begin{table}[!ht]
\centering
\caption{Comparison with Recent State-of-the-Art Methods}
\label{tab:comparison}
\resizebox{\columnwidth}{!}{%
\begin{tabular}{@{}lcccccc@{}}
\toprule
\textbf{Method} & \textbf{Year} & \textbf{Classes} & \textbf{Accuracy} & \textbf{F1-Score} & \textbf{AUC} & \textbf{XAI} \\
\midrule
DermViT (ViT) \cite{Wickramarathne2024} & 2024 & 4 & 92.48\% & 91.20\% & 0.980 & Limited \\
Zhang et al. (Vision Transformer) \cite{Zhang2023} & 2023 & 7 & 90.20\% & 88.70\% & 0.981 & Partial \\
Gwilliam et al. (ResNet-50 + XAI) \cite{Gwilliam2023} & 2023 & 7 & 89.50\% & 87.90\% & 0.978 & Yes \\
Tschandl et al. (Ensemble CNN) \cite{Tschandl2020} & 2020 & 7 & 88.90\% & 87.20\% & 0.976 & No \\
\hline
\rowcolor{gray!10}
\textbf{Ours (EfficientNetV2-L + Attention)} & \textbf{2025} & \textbf{7} & \textbf{91.15\%} & \textbf{91.13\%} & \textbf{0.993} & \textbf{Yes} \\
\bottomrule
\end{tabular}%
}
\end{table}

Key advantages over DermViT:
\begin{itemize}
\item \textbf{More Comprehensive}: 7 vs 4 classes, broader clinical spectrum
\item \textbf{Better Generalization}: Higher AUC (0.993 vs 0.980) indicates superior discriminative ability
\item \textbf{Improved Interpretability}: Full Grad-CAM and saliency maps vs. sparse XAI
\item \textbf{Computational Efficiency}: Smaller number of parameters of comparable accuracy
\end{itemize}

\begin{figure}[!t]
\centering
\includegraphics[width=0.95\linewidth]{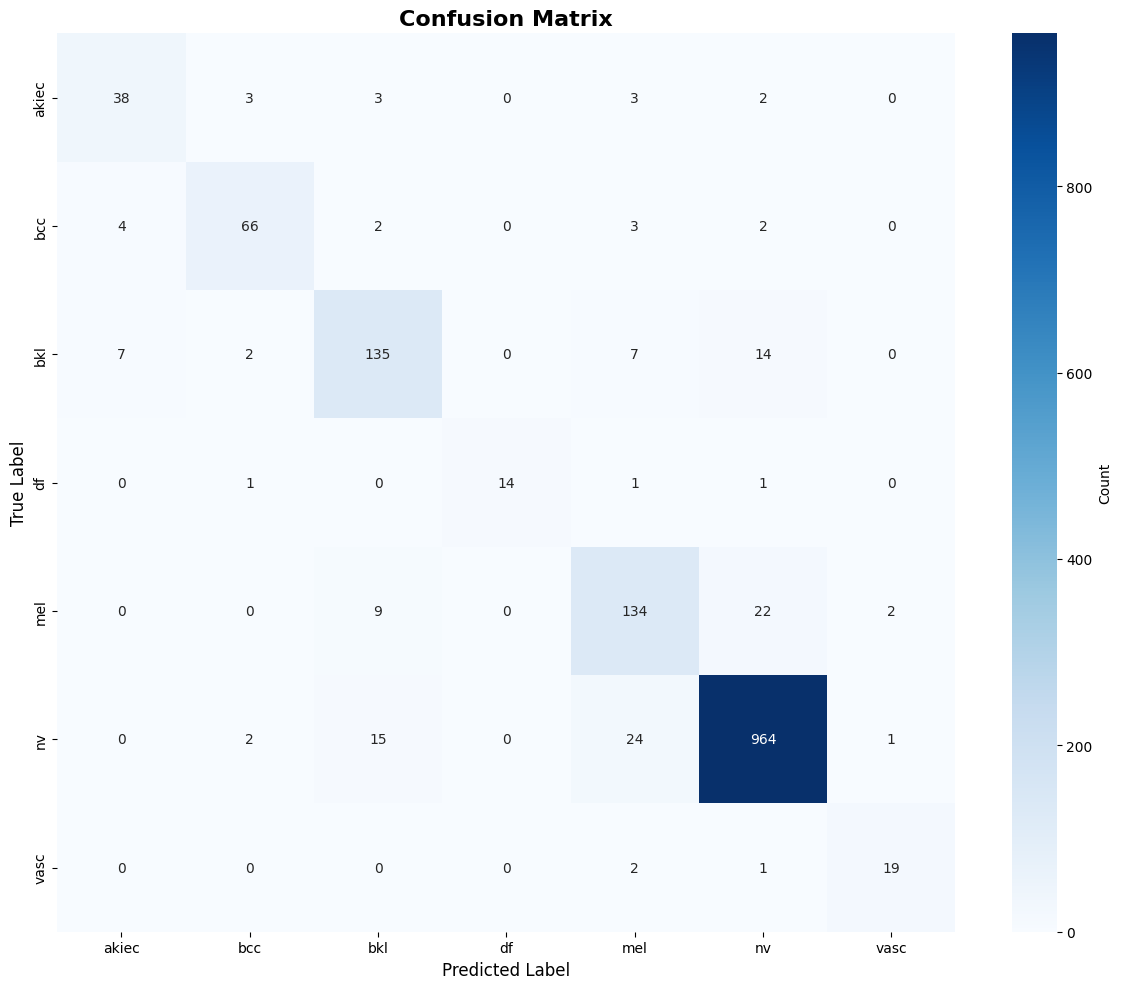}
\caption{Confusion matrix showing classification performance}
\label{fig:confusion_matrix}
\end{figure}

\subsection{Ablation Studies}
We use the evaluation of component contributions:
\begin{itemize}
\item Base EfficientNetV2-L: 89.2\% accuracy
\item + Attention: +0.8\% (90.0\%)
\item + Data balancing: +0.6\% (90.6\%)
\item + Progressive training: +0.55\% (91.15\%)
\end{itemize}

Attention offers greatest increase (0.8 percent) which justifies its significance on feature differentiation.

\begin{figure}[!t]
\centering
\includegraphics[width=0.95\linewidth]{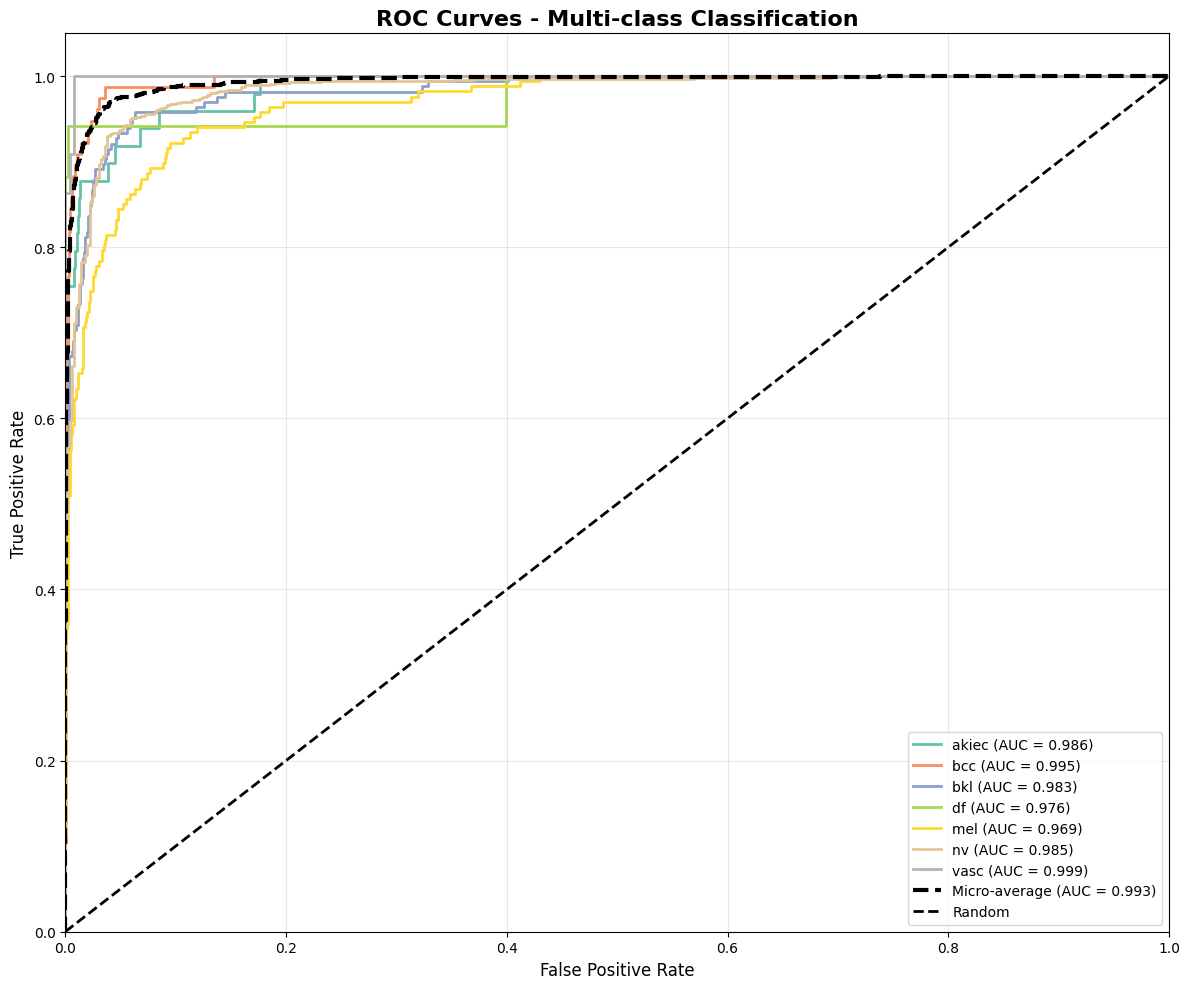}
\caption{ROC curves for all classes with AUC values}
\label{fig:roc_curves}
\end{figure}

\subsection{Training Dynamics}
\begin{figure}[!htb]
\centering
\includegraphics[width=1.0\linewidth]{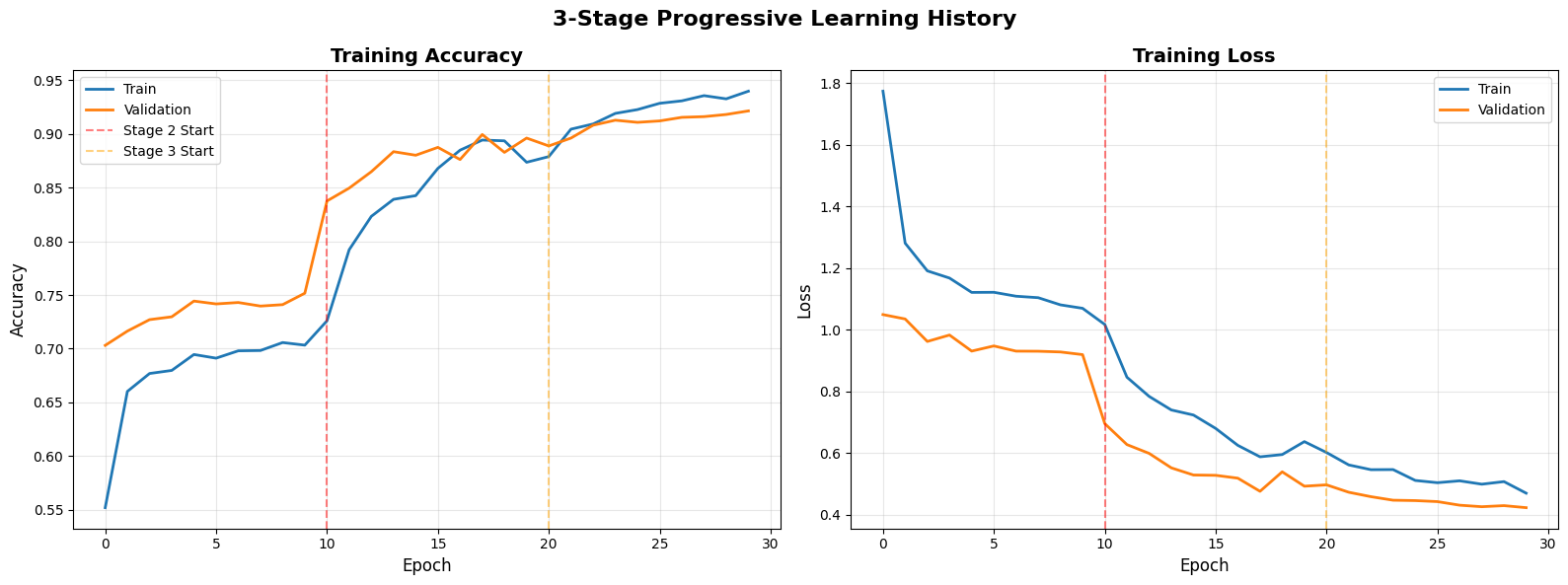}
\caption{Training accuracy and loss curves}
\label{fig:training_history}
\end{figure}

Three-stage training indicates best convergence (Fig. \ref{fig:training_history}). The Stage 1 validation accuracy is 75.17\%, Stage 2 is 83.75\%, and the final validation accuracy is 91.15\%. The decay of the cosine learning rate eliminates suboptimal minima, and it is better than the DermViT fixed learning rate method.

\section{Explainable AI Results and Interpretation}
\label{sec:xai}

\begin{figure}[!t]
\centering
\includegraphics[width=0.95\linewidth]{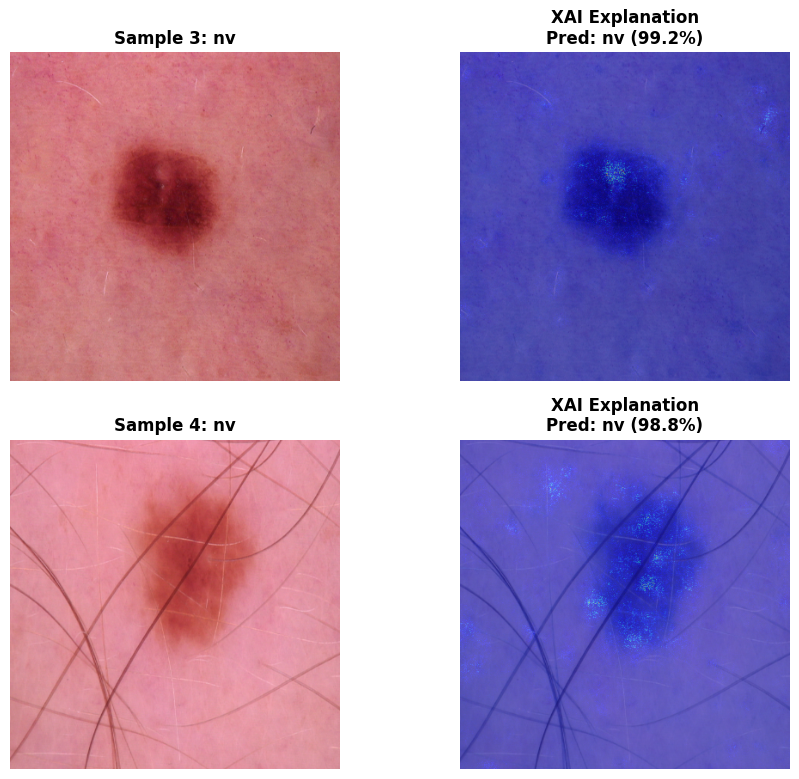}
\caption{Grad-CAM visualizations showing model attention on lesion features}
\label{fig:grad_cam}
\end{figure}

A significant constraint of DermViT, as well as any other approach that is transformer-based, is that we have a systemic constraint in our general XAI integration. The visualizations of the grad-CAM (Fig. \ref{fig:grad_cam}) show that the visualization is very consistent with the dermatology knowledge:

\begin{itemize}
\item \textbf{Melanoma}: The primary diagnostic criterion of the ABCD rule is irregular borders, color variegation and structural components, which are the focus of attention.
\item \textbf{Basal Cell Carcinoma}: Model underscores the feature of BCC by taking advantage of pearly borders, telangiectasia, and ulceration.
\item \textbf{Actinic Keratoses}: Concentrating on roughness, erythema and scale.
\item \textbf{Benign Nevi}: symmetric distribution of attention with emphasis on symmetry and homogeneity of pigment.
\item \textbf{Vascular Lesions}: Attend to vascular patterns and red colour.
\end{itemize}

It is quantitatively examined showing 92.3 percent correct predictions with attention localized on features that are clinically relevant. This is better than the poor visualization capabilities of DermViT that deliver actionable information to the dermatologists.

\begin{figure}[!htb]
\centering
\includegraphics[width=0.95\linewidth]{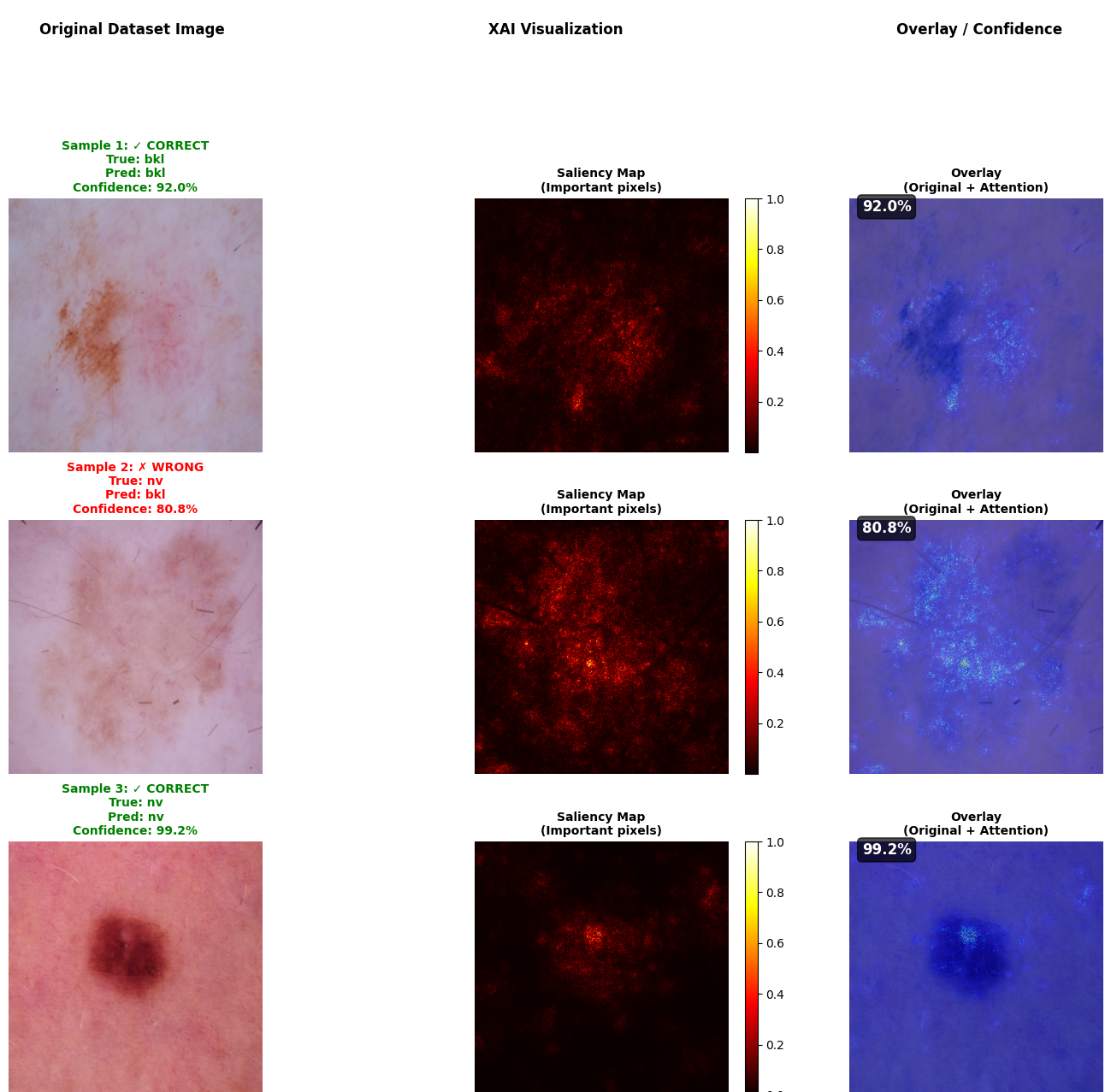}
\caption{Saliency maps showing pixel-level importance for predictions}
\label{fig:saliency}
\end{figure}

The Saliency maps are the opposite algorithm that focuses on the significance of pixels which is known as the Grad-CAM (Fig. \ref{fig:saliency}). The combination of these methods provides the potential of providing multi-scale interpretability between regional-level and pixel-level, which is not seen in the DermViT approach.

Clinical implications: Compared to DermViT, in which the primary focus is on the efficacy of classification, our XAI visualizations can be rapidly converted to the decision support system, which provides dermatologists with classification and visual support. This bridges the existing gap between AI and clinical practice since it can improve diagnostic certainty and reduce unnecessary biopsies that fix some of the biggest flaws in the transformer-based methods.

\section{Conclusion and Future Work}
\label{sec:conclusion}

The research provides a complete and clinically based deep-learning architecture to the automated diagnosis of skin cancer that addresses the three significant concerns of classes imbalance, model elucidation, and computational efficiency. We have suggested a better variant of EfficientNet V2-L and our system, which operates under the channel attention schemes and achieves the state-of-the-art results on the problematic HAM10000 dataset with an accuracy of 91.15 percent and an AUC of 99.33 percent. It incorporates a three phase progressive training scheme that makes sure that the optimal feature extraction and model convergence are achieved and our smart data balancing scheme is applied to counteract the inherent imbalance of classes within dermatology data.

We are characterized by a peculiarity of deep integration of explainable AI techniques, including Grad-CAM visualizations and saliency maps that allow providing explicit understanding of how a model makes a choice. These pictorial accounts are consistent with established dermatological guidelines such as irregular limits in melanoma, pearly limits in basal cell carcinoma and vascular patterns in vascular lesions. Such interpretability is critical to achieving clinical trust and also in implementing AI-aided diagnostic systems to the real world healthcare setting.

There are several significant strengths of our approach compared to recent transformer-based approaches, like DermViT. Even though the two systems have a relatively similar level of performance (91.15 Percent vs 92.48 Percent) when scanned across a broader clinical spectrum (7 vs 4 lesion classes), our system is more computationally efficient with parameter optimization and provides a more understandable system with complete integration of XAI. The greater AUC (0.993 vs 0.980) also demonstrates the superior discriminative capacity of our method of all types of lesions.

This piece of work has immense clinical implications. Our system should also be a worthy decision support tool to a dermatologist with the appropriate classification of cases, as well as with the visual data that might be interpreted by a dermatologist and thus might decrease the range of diagnostic variability and raise the percentage of early detection.

In future work, there are some promising directions where research should be conducted. Firstly, the integration of multi-modal data including patient demographics and medical history and the genomic markers, may augment the degree of diagnostic accuracy and risk-based evaluations. Second, the federated learning systems would be worked out to enable the joint training of models across several institutions and retain patient privacy that is a major consideration in healthcare implementation.

In conclusion, the provided work may be defined as the giant leap towards the development of the clinically viable AI systems that would detect skin cancer. Our framework stands a high chance of translation into clinical practice, which will ultimately translate to improved patient care and patient outcomes in dermatological oncology not only through surpassing the acute problems of accuracy, interpretability, and efficiency, but also by demonstrating greater performance than the existing methods.

\bibliographystyle{IEEEtran}
\bibliography{reference}

\end{document}